%% file: main.tex
\documentclass[10pt]{article}
\usepackage[accepted]{tmlr}

\input{math_commands.tex}

\usepackage{hyperref}
\usepackage{url}
\usepackage{amsmath}
\usepackage{amssymb}
\usepackage{amsthm}

\input{macros}

\theoremstyle{plain}
\newtheorem{theorem}{Theorem}[section]

\newtheorem{lemma}[theorem]{Lemma}

\theoremstyle{definition}

\theoremstyle{remark}

\title{On the Robustness of Dataset Inference}

\author{\name Sebastian Szyller \email contact@sebszyller.com \\
      \addr Aalto University
      \AND
      \name Rui Zhang \email zhangrui98@zju.edu.cn \\
      \addr Zhejiang University
      \AND
      \name Jian Liu\email liujian2411@zju.edu.cn\\
      \addr Zhejiang University
      \AND
      \name N. Asokan \email asokan@acm.org \\
      \addr University of Waterloo \& Aalto University}

\def\month{06}  
\def\year{2023} 
\def\openreview{\url{https://openreview.net/forum?id=LKz5SqIXPJ}} 

\begin{document}

\maketitle
\input{0abstract}
\input{1introduction}
\input{2background}
\input{3falsepositives}
\input{4falsenegatives}
\input{5countermeasures}
\input{6discussion}
\input{8conclusion}

\section*{Acknowledgement}
We would like to thank the authors of Dataset Inference~\citep{maini2021datasetinference} for their feedback on our work presented in this paper. This work was supported in part by Intel (in the context of the Private-AI Institute),
National Natural Science Foundation of China (Grant No. 62002319, U20A20222),
Hangzhou Leading Innovation and Entrepreneurship Team (TD2020003),
and
Zhejiang Key R\&D Plan (Grant No. 2021C01116).

\bibliography{bibliography}
\bibliographystyle{tmlr}

\input{9appendix}

\end{document}

%% file: math_commands.tex
\usepackage{amsmath,amsfonts,bm}

\def\eqref#1{equation~\ref{#1}}

\def\1{\bm{1}}

\def\vc{{\bm{c}}}

\def\vx{{\bm{x}}}

\DeclareMathAlphabet{\mathsfit}{\encodingdefault}{\sfdefault}{m}{sl}
\SetMathAlphabet{\mathsfit}{bold}{\encodingdefault}{\sfdefault}{bx}{n}



%% file: macros.tex
\usepackage{amsmath,amsfonts}
\usepackage{dsfont}
\usepackage{graphicx}
\usepackage{textcomp}
\usepackage{xcolor}
\usepackage{colortbl}
\usepackage{xspace}
\usepackage{tabularx}
\usepackage{cleveref}
\usepackage{multirow}
\usepackage{subcaption}
\usepackage[inline]{enumitem}
\usepackage{soul}
\usepackage{xurl}
\usepackage{wrapfig}
\usepackage{float}
\usepackage{subcaption}

\newcolumntype{Y}{>{\centering\arraybackslash}X}

\newcommand{\adv}{\ensuremath{\mathcal{A}}}
\newcommand{\victim}{\ensuremath{\mathcal{V}}}
\newcommand{\independent}{\ensuremath{\mathcal{I}}}

\newcommand{\fv}{\ensuremath{f_{\mathcal{V}}}}
\newcommand{\fa}{\ensuremath{f_{\mathcal{A}}}}
\newcommand{\fo}{\ensuremath{f_{0}}}
\newcommand{\find}{\ensuremath{f_{\mathcal{I}}}}
\newcommand{\fsp}{\ensuremath{f_{SP}}}
\newcommand{\distinguisher}{\ensuremath{g_{\mathcal{V}}}}

\newcommand{\distribution}{\ensuremath{\mathcal{D}}}
\newcommand{\dataset}{\ensuremath{\mathcal{S}}}
\newcommand{\privatedata}{\ensuremath{\mathcal{S}_{V}}}
\newcommand{\publicdata}{\ensuremath{\mathcal{S}_{0}}}
\newcommand{\independentdata}{\ensuremath{\mathcal{S}_{I}}}

\newcommand{\di}{\textsf{DI}}

\newcommand{\bad}[1]{\textcolor{red}{\underline{#1}}}
\newcommand{\techreport}[1]{}

\newcommand{\oldtext}[2]{}

%% file: 0abstract.tex

\begin{abstract}
Machine learning (ML) models are costly to train as they can require a significant amount of data, computational resources and technical expertise.
Thus, they constitute valuable intellectual property that needs protection from adversaries wanting to steal them.
\emph{Ownership verification} techniques allow the victims of model stealing attacks to demonstrate that a suspect model was in fact stolen from theirs.

Although a number of ownership verification techniques based on watermarking or fingerprinting have been proposed, most of them fall short either in terms of security guarantees (well-equipped adversaries can evade verification) 
or computational cost.
A fingerprinting technique, \emph{Dataset Inference} ($\di$), 
has been shown to offer better robustness and efficiency than prior methods.

The authors of $\di$ provided a correctness proof for linear (suspect) models.
However, in a subspace of the same setting, we prove that $\di$ suffers from high false positives (FPs) -- it can incorrectly identify an independent model trained with non-overlapping data from the same distribution as stolen.
We further prove that $\di$ also triggers FPs in realistic, non-linear suspect models.
We then confirm empirically that $\di$ in the black-box setting leads to FPs, with high confidence.


Second, we show that $\di$ also suffers from false negatives (FNs) -- an adversary can fool $\di$ (at the cost of incurring some accuracy loss) by regularising a stolen model's decision boundaries using adversarial training, thereby leading to an FN.
To this end, we demonstrate that black-box $\di$ fails to identify a model adversarially trained from a stolen dataset -- the setting where $\di$ is the hardest to evade.

Finally, we discuss the implications of our findings, the viability of fingerprinting-based 
ownership verification in general, and suggest directions for future work.
\end{abstract}

%% file: 1introduction.tex

\section{Introduction}
\label{sec:introduction}

Machine learning (ML) models are being developed and deployed at an increasingly faster rate and in several application domains.
For many companies, they are not just a part of the technological stack that offers an edge over the competitors but a core business offering.
Hence, ML models constitute valuable intellectual property that needs to be protected.

Model stealing is considered one of the most serious attack vectors against ML models~\citep{microsoft:2019}.
The goal of a model stealing attack is to obtain a functionally equivalent copy of a victim model that can be used, for example, to offer a competing service, or avoid having to pay for the use of the model.

In the \emph{white-box} attack, the adversary obtains the exact copy of the victim model, for example by reverse engineering an application containing an embedded model~\citep{meng2022appsml}.
In contrast, in \emph{black-box} attacks (known as \emph{model extraction} attacks)~\citep{papernot2017practical,orekondy2018knockoff,tramer2016stealing} 
the adversary gleans information about the victim model via its predictive interface.
Two possible approaches to defend against model extraction are 1) detection~\citep{juuti2019prada,atli2020boogeyman,zheng2020blp} and 2) prevention~\citep{orekondy20prediction,mazeika2022steeringnoise,dziedzic2022increasing}.
However, a powerful, yet realistic attacker can circumvent these defenses~\citep{atli2020boogeyman}.


An alternative defense applicable to both white-box and black-box model theft is based on \emph{deterrence}.
It concedes that the model will eventually get stolen.
Therefore, an \emph{ownership verification} technique that can identify and demonstrate a suspect model as having been stolen can serve as a deterrent against model theft.
Early research in this field focused on \emph{watermarking} based on embedding triggers or backdoors~\citep{zhang2018protecting,uchida2017embedding,adi2018turning} into the weights of the model.
Unfortunately, all watermarking schemes were shown to be brittle~\citep{lukas2021wmsok} in that an attacker can successfully remove the watermark from a protected stolen model without incurring a substantial loss in model utility.

An alternative approach to ownership verification is \emph{fingerprinting}.
Instead of embedding a trigger or backdoor in the model, one can extract a fingerprint that matches only the victim model, and models derived from it.
Fingerprinting works both against white-box and black-box attacks, and does not affect the performance of the model.
Although several fingerprinting schemes have been proposed, some are not rigorously tested against model extraction~\citep{cao2021ipguard,pan2022metav} and others can be computationally expensive to derive~\citep{lukas2021conferrable}.

In this backdrop, \emph{Dataset Inference} ($\di$)~\citep{maini2021datasetinference} promises to be an effective fingerprinting mechanism.
Intuitively, it leverages the fact that if model owners trained their models on \emph{private data}, knowledge about that data can be used to identify all stolen models.
$\di$ was shown to be effective against white-box and black-box attacks and is efficient to compute~\citep{maini2021datasetinference}.
It was also shown not to conflict with any other defenses~\citep{szyller22conflicts}.
Given its promise, the guarantees provided by $\di$ merits closer examination.

In this work, we first show that $\di$ suffers from false positives (FPs) --- it can incorrectly identify an independent model trained with \emph{non-overlapping data from the same distribution} as stolen.
The authors of $\di$ provided a correctness proof for a linear model.
However, $\di$ in fact suffers from \textbf{high FPs}, unless two assumptions hold: (1) a large noise dimension, as explained in the original paper and (2) a large proportion of the victim's training data is used during ownership verification, as we prove in this paper. Both of these assumptions are unrealistic in a subspace of the linear case used by $\di$: (i) we prove that a large noise dimension can lead to low accuracy in the resulting model , and (ii) revealing too much of the victim's (private) training data is detrimental to privacy.
Furthermore, we prove that $\di$ also triggers FPs in realistic, non-linear models.
We then confirm empirically that $\di$ leads to FPs, with high confidence in the black-box verification setting, ``\emph{black-box} $\di$'', where the $\di$ verifier has access only to the inference interface of a suspect model, but not its internals .

We also show that black-box $\di$ suffers from false negatives (FNs):
an adversary who has in fact stolen a victim model can avoid detection by regularising their model with adversarial training.
We provide empirical evidence that an adversary who steals the victim's dataset itself and adversarially trains a model can evade detection by $\di$ by trading off accuracy of the stolen model.
 
We claim the following contributions:
\begin{itemize}
    \item Following the same simplified theoretical analysis used by the original paper~\citep{maini2021datasetinference}, in a subspace of the linear case used by $\di$, we show that for a linear suspect model, 1) high-dimensional noise (as required in~\citep{maini2021datasetinference} leads to \textbf{low model accuracy} (Lemma \ref{lemma.1}, Section \ref{sec:fps-linear}), and 2) $\di$ \textbf{suffers from FPs} unless a large proportion of private data is revealed during ownership verification (Theorem~\ref{lemma.2}, Section \ref{sec:fps-linear});
    \item Extending the analysis to non-linear suspect models, using a PAC-Bayesian framework~\citep{neyshabur2018pac}, we show that $\di$ suffers from \textbf{FPs in non-linear models} regardless of how much private data is revealed 
    (Theorem~\ref{thm.2}, Section \ref{sec:nl-theory});
    \item We empirically demonstrate  the existence of \textbf{FPs in a realistic black-box $\di$} setting (Section \ref{sec:nl-evidence});
    \item We show empirically that black-box $\di$ also \textbf{suffers from FNs}: using adversarial training to regularise the decision boundaries of a stolen model can successfully evade detection by $\di$ at the cost of some accuracy  
    ($\approx$ 6-13pp) (Section \ref{sec:false-negatives});
\end{itemize}

%% file: 2background.tex

\section{Dataset Inference Preliminaries}
\label{sec:background}

Dataset Inference ($\di$) aims to determine whether a \emph{suspect model} $\fsp$ was obtained by an adversary $\adv$ who has stolen a model ($\fa$) derived from a victim $\victim$'s private data $\privatedata$, or belongs to an independent party $\independent$ ($\find$).
$\di$ relies on the intuition that if a model is derived from $\privatedata$, this information can be identified from all models.
$\di$ measures the \emph{prediction margin}s of a suspect model around private and public samples: distance from the samples to the model's decision boundaries.
If $\fsp$ has distinguishable decision boundaries for private and public samples $\di$ deems it to be \emph{stolen}; otherwise the model is deemed \emph{independent}.

In the rest of this section, we explain the theoretical framework that $\di$ uses --- consisting of a linear suspect model --- the embedding generation necessary for using $\di$ with realistic non-linear suspect models, and the verification procedure.
A summary of the notation used throughout this work appears in Table~\ref{tab:notations}.

\subsection{Theoretical Framework}
\label{sec:bg-theoretical}

The original $\di$ paper~\citep{maini2021datasetinference} used a linear suspect model to theoretically prove the guarantees provided by $\di$.
We first explain how $\di$ works in this setting.

\textbf{Setup.} Consider a data distribution $\distribution$, such that any input-label pair $(\bm{x}, y)$ can be described as:
\begin{equation*}
    y\sim\{-1,+1\},\bm{x_1}=y\cdot \bm{u} \in \mathbb{R}^K, \bm{\bm{x_2}}\sim \mathcal{N}(0,\sigma^2I)\in\mathbb{R}^D,
\end{equation*}
where $\bm{x}=(\bm{x_1},\bm{\bm{x_2}})\in\mathbb{R}^{K+D}$ and $\bm{u}\in\mathbb{R}^K$ is a fixed vector. The last $D$ dimensions of $\bm{x}$ represent Gaussian noise (with variance $\sigma^2$).

\textbf{Structure of the linear model.}
Assuming a linear model $f$, with weights $\bm{w}=(\bm{w_1},\bm{w_2})$, such that $ f(x)=\bm{w_1}\cdot \bm{x_1}+\bm{w_2}\cdot \bm{x_2}$, then the final classification decision is $sgn(f(x))$.
With the weights initialized to zero, $f$ learns the weights using gradient descent with learning rate $1$ until $yf(x)$ is maximized.
Given a private training dataset $\privatedata \sim \distribution = \{(x^{(i)},y^{(i)})| i=1,...,m\}$, and a public dataset $\publicdata\sim \distribution$ (both of size $m$),
then $\bm{w_1} = m\bm{u}$ and $\bm{w_2} = \sum_{i=1}^{m} y^{(i)}\bm{x_2}^{(i)}$ regardless of the batch size.

In $\di$, the prediction margin $p(\cdot)$ is used to imply the confidence of $f$ in its prediction.
It is defined as the margin (distance) of a data point from the decision boundary.
\begin{equation}\label{eq:margins}
    p(x) \triangleq y\cdot f(x).
\end{equation}



The authors~\citep{maini2021datasetinference} show that the difference of expected prediction margins of two datasets $\privatedata$ and $\publicdata$ is $D\sigma^2$.
The threshold can be set $\lambda\in(0, D\sigma^2)$, and by estimating the difference of the prediction margins on $\publicdata$ and $\privatedata$ on $\fsp$, $\di$ is able to distinguish whether that model is stolen.

Note that $\di$ uses approximations of the prediction margins based on embeddings.
The theoretical framework assumes that the approximations are accurate, and we can use them directly for the theoretical analysis (Equation~\ref{eq:margins}).
For the linear model, the margins can be computed analytically;
however, in Section~\ref{sec:bg-practice}, we explain how the approximations of the margins are obtained.

\subsection{Embedding Generation}
\label{sec:bg-practice}

In order to use $\di$ one needs to generate \emph{embeddings} of the samples.
$\victim$ queries their model $\fv$ with samples in their private dataset $\privatedata$ and public dataset $\publicdata$, and assigns the labels $b=1$ and $b=0$ respectively.
The authors propose two methods of generating the embeddings: a white-box approach (MinGD) and a black-box one (Blind Walk).
In this work, we use only Blind Walk as it outperforms MinGD in most experimental setups in the original work, and is more realistic, as it only requires access to the API of the suspect model.

Blind Walk estimates the prediction margin of a sample by measuring its robustness to random noise.
For a sample $(\vx, y)$, to compute the margin, first choose a random direction $\delta$, and take $k \in \mathbb{N}$ steps in the same direction until the misclassification $f(\vx + k\delta) \neq y$.
This is repeated multiple times to increase the size of the embedding.
As reported in \citep{maini2021datasetinference}, obtaining embeddings for $100$ samples can take up to $30,000$ queries.

Having obtained the embeddings, $\victim$ trains a regression model $\distinguisher$ that predicts the confidence that a sample contains private information from $\privatedata$.

\begin{table}[t]
\caption{Summary of the notation used throughout this work.}
\centering
\begin{tabular}{l|l|l|l}
    \hline
    $\victim$          & the victim                            & $f$                & a model\\
    $\independent$     & an independent party                  & $\fv$              & a model trained on $\privatedata$\\
    $\adv$             & an adversary                          & $\fo$              & a model trained on $\publicdata$\\
    $\dataset$         & a dataset                             & $\find$            & a model trained on $\independentdata$\\
    $\privatedata$     & $\victim$'s private dataset           & $\fa$              & $\adv$'s model     \\
    $\publicdata$      & a public dataset                      & $\fsp$             & a suspect model      \\
    $\independentdata$ & $\independent$'s data                 & $\bm{w}$           & model weights      \\
    $\distribution$    & distribution that all datasets follow & $\distinguisher$   & regression model \\
    $(\bm{x},y)$       & a sample from $\distribution$         & $D$                & noise dimension          \\
    \hline
\end{tabular}\label{tab:notations}
\end{table}

\subsection{Ownership Verification}
\label{sec:bg-verification}

Using the scores from $\distinguisher$ and the membership labels, $\victim$ creates vectors $\vc$ and $\vc_{V}$ of equal size from $\privatedata$ and $\publicdata$, respectively.
Then for a null hypothesis $H_0 : \mu < \mu_{V}$ where $\mu = \Bar{c}$ and $\Bar{\mu} = \Bar{c_{V}}$ are mean confidence scores.
The test rejects $H_0$ and rules that the suspect model is ‘stolen’, or gives an inconclusive result.

\begin{wrapfigure}[16]{r}{0.5\textwidth}
    \centering
      \includegraphics[width=0.45\textwidth]{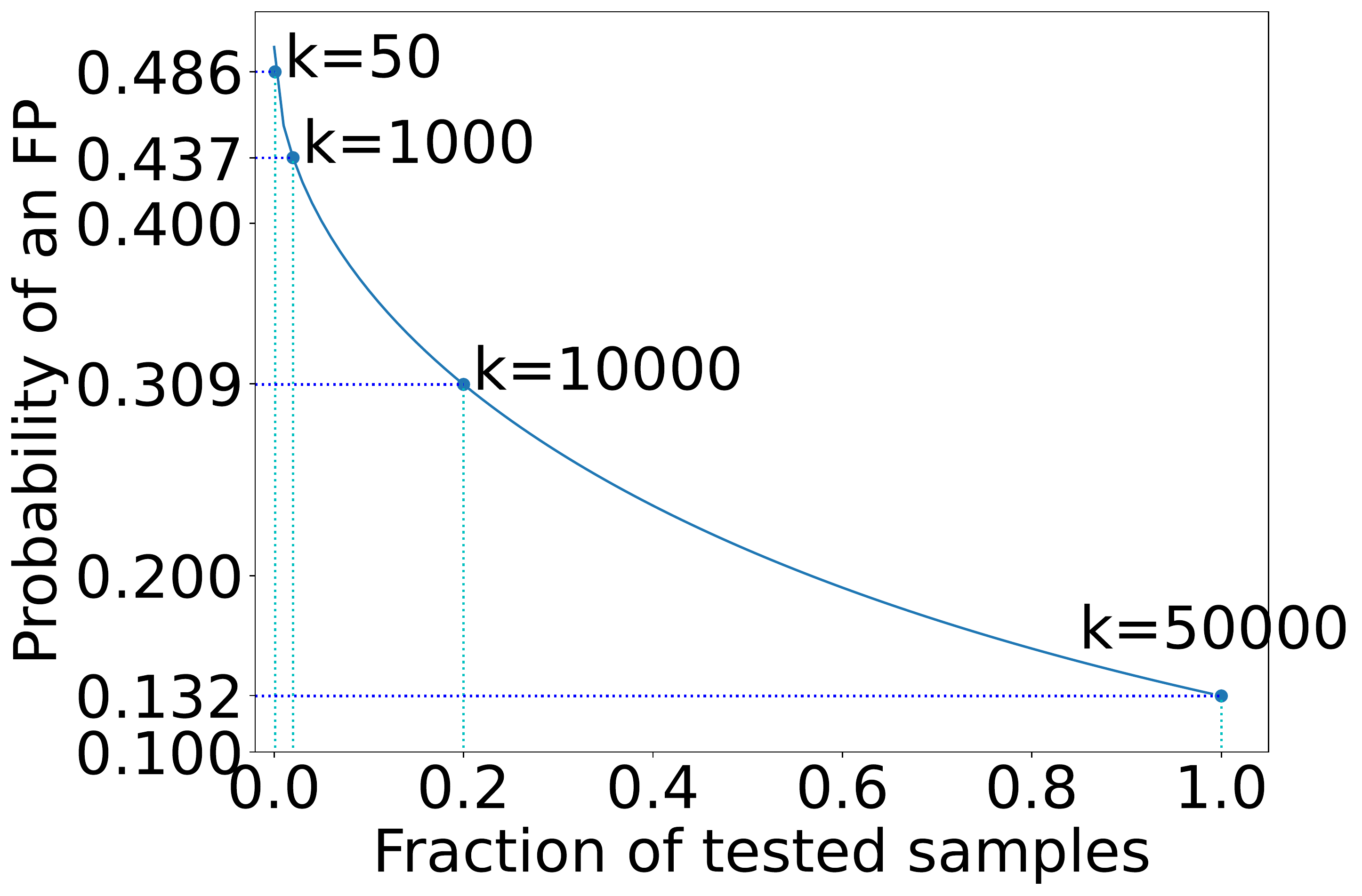}
    \caption{Probability of an FP as the fraction of revealed private samples for $D=10$ for a linear suspect model (Equation~\ref{eq.fp}). $\victim$ needs to use many private samples to guarantee a low false positive rate.}
    \label{fig:fpr}
\end{wrapfigure}

To verify whether $\fsp$ is stolen or independent, $\victim$ obtains the embeddings by querying the model (using Blind Walk) using samples from $\privatedata$ and $\publicdata$.
Then they use the embeddings to obtain the confidence scores from the $\distinguisher$, and perform a hypothesis test on the two distributions of scores.

%% file: 3falsepositives.tex

\section{False Positives in Dataset Inference}
\label{sec:false-positives}

To generate the embeddings for a specific sample in the private dataset $\privatedata$, $\di$ requires querying the suspect model $\fsp$ hundreds of times.
To reduce the total number of queries, $\di$ was shown to be effective with only $10$ private samples with at least $95\%$ confidence.
Additionally, $\di$ requires a large random noise dimension $D$ such that probability of success increases to $1$ as $D \rightarrow \infty$.
In this section, we prove that these two assumptions are not realistic in a subspace of the linear case used by $\di$:
1) $\di$ is susceptible to false positives (FPs) unless $\victim$ reveals a large number of samples;
2) a large $D$ will harm the utility of the model (Section~\ref{sec:fps-linear}).

Furthermore, we find that the theoretical results on linear suspect models which say that the margins on different models are distinguishable with some strict conditions do not hold for more realistic non-linear suspect models.
Using a PAC-Bayesian margin based generalization bound~\citep{neyshabur2018pac} we prove that models trained on the same distribution are indistinguishable, and will trigger FPs (Section~\ref{sec:nl-theory}.
Next, we provide empirical evidence for the existence of FPs (Section~\ref{sec:nl-evidence}).

\subsection{Linear Suspect Models}
\label{sec:fps-linear}

In Section~\ref{sec:background}, we have a set up for linear models.
$\di$ shows their success theoretically on the linear set up where $\bm{u}\in\mathbb{R}^K$ and $\sigma\in \mathbb{R}$.

\begin{theorem}[Success of DI~\citep{maini2021datasetinference}]
Choose $b\leftarrow \{0,1\}$ uniformly at random. Given an adversary's linear classifier $f$ trained on $\dataset \sim \distribution, s.t. |S|=\privatedata$ if $b=0$, and on $\privatedata$ otherwise. The probability $\victim$ correctly decides if an adversary stole its knowledge $\mathbb{P}[\Psi(f,\privatedata;\distribution)=b]=1-\Phi(-\frac{\sqrt{D}}{2\sqrt{2}}).$ Moreover, $lim_{D\rightarrow \infty}\mathbb{P}[\Psi(f,\privatedata;\distribution)=b]=1.$
\end{theorem}

However, realistic datasets often contain more noise and lack the correct signal.
We consider a subspace where the signal $||\bm{u}||$ is upper-bounded and the noise is larger.
We construct the subspace by assuming that $||\bm{u}||_2\leq\frac{1}{\sqrt{m}}$ and $\sigma^2>\frac{1}{10\sqrt{m}}$.
Note that the linear model should correctly classify most of the randomly picked data from this subspace, we use Lemma~\ref{lemma.2} to analyze the accuracy of the linear model in this subspace.
The results show that only the subspace where $D$ is bounded can guarantee high performance for the model.



\begin{lemma}[Need for Bounding Noise Dimension]\label{lemma.2}
Let $f$ be a linear model trained on $\dataset \sim \distribution$.
For a sample $(\bm{x},y)$ sampled from $\distribution$ which is independent of $\dataset$, $f$ correctly classifies the sample with a probability $\mathbb{P}[yf(x)\geq0]=1-\Phi(-\frac{m\bm{u}^2}{\sqrt{mD}\sigma^2})$.
Assuming that $||\bm{u}||_2 \leq \frac{1}{\sqrt{m}}$ and $\sigma^2 > \frac{1}{10\sqrt{m}}$, the accuracy of $f$ is $\mathbb{P}[yf(x)\geq0]\leq1-\Phi(-\frac{10}{\sqrt{D}})$. Therefore, the model $f$ trained on the $\mathcal{D}$ can achieve high accuracy only if $D$ is bounded.
\end{lemma}
The details of the proof are in the Appendix~\ref{app:lem.1}.
Lemma~\ref{lemma.2} shows that if the dimension of $\bm{x}_2$, which follows $\mathcal{N}(0,\sigma^2)$, is large, then the noise will dominate $f$ and mislead it into making incorrect predictions.
For example,
for a dataset with more than 500 samples, the std of $\bm{x_2}$ is $0.25$ (close to the CIFAR10 dataset) where $\sigma^2=0.0625 > \frac{1}{10\sqrt{m}}=\frac{1}{10\sqrt{500}}\approx0.0044$. If $D=1000$, $f$ can correctly classify a sample that is different from $f$'s training set with a probability up to $1-\Phi(-\frac{10}{\sqrt{1000}})=0.6241$.
However, if $D=10$, $f$ can achieve an accuracy of 0.9992.

Since $\victim$ limits the number of samples used for the verification, Theorem~\ref{lemma.1} show that the false positives (FPs) rate is directly related to the samples used for the verification in our subspace.

\begin{theorem}[Existence of False Positives with Linear Suspect Models]\label{lemma.1}
Let $\find$ be a linear classifier trained on the independent dataset $\independentdata \sim \distribution$.
Let $k$ be the number of samples estimated required for the verification.
Then,
the probability that $\victim$ mistakenly decides that $\find$ is a stolen model is $\mathbb{P} [\Psi(\find, \privatedata; \distribution) = 1] > 1 -\Phi(\frac{\sqrt{kD}}{2\sqrt{2m}})$.
\end{theorem}

Where $\Psi$ is $\victim$'s decision function~\citep{maini2021datasetinference}:
\begin{equation}
\Psi(\fsp, \dataset; \distribution) =   \left\{
\begin{aligned}
&1,\ if\ \fsp\sim\fa, \\
&0,\ if\ \fsp\sim\find,
\end{aligned}
\right.
\end{equation}


\begin{proof}
Recall that $\victim$ tries to reveal only a few samples during the verification.
For a distribution $\distribution$ where $||\bm{u}||\leq\frac{1}{\sqrt{m}}$ and $\sigma^2>\frac{1}{\sqrt{m}}$.

Following the intuition from $\di$~\citep{Yeom}, for satisfactory performance, $\di$ must minimise both false positives and false negatives.
Hence, the objective function is defined as:
\begin{equation}
    min_\lambda \frac{\mathbb{P}[\Psi(\find, \privatedata; \distribution) = 1]+\mathbb{P}[\Psi(\fv, \privatedata; \distribution) = 0]}{2},
\end{equation}
where the margin of $\distribution$ is estimated using $\privatedata$ and $\publicdata$.
Note that we are only interested in the false positives $\mathbb{P}[\Psi(\find, \privatedata; \distribution) = 1]$, let $\independentdata = \{(x^{(i)},y^{(i)})| i=1,...,m\}$, $\dataset_*^k$ be a subset of $\dataset_*$ consisting of $k$ samples.

\begin{equation}\label{equ.term}
\begin{aligned}
    &\quad\ \mathbb{P}[\Psi(\find, \privatedata; \distribution) = 1]
    \\&=\mathbb{P} [E_{(\bm{x},y)\in\privatedata^k}[y \find(\bm{x})]- E_{(\bm{x},y)\in\publicdata^k}[y \find(\bm{x})]\geq \lambda]
    \\&=\mathbb{P} [E_{(\bm{x},y)\in\privatedata^k}[\sum_i^m y^{(i)}\bm{x_2}^{(i)}\bm{x_2}]
    \ - E_{(\bm{x},y)\in\publicdata^k}[\sum_i^m y^{(i)}\bm{x_2}^{(i)}\bm{x_2}]\geq\lambda]
    \\&=\mathbb{P} [\frac{1}{k}\sum_j^k \sum_i^m y^{(i)}\bm{x_2}^{(i)}\bm{x_2}^{(j)} - \frac{1}{k}\sum_p^k \sum_i^m y^{(i)}\bm{x_2}^{(i)}\bm{x_2}^{(p)} \geq\lambda].
\end{aligned}
\end{equation}
Recall that $\bm{x_2}^{(i)}$, $\bm{x_2}^{(j)}$ and $\bm{x_2}^{(p)}$ are $D$-dimensional vectors sampled independently from $\mathcal{N}(0,\sigma^2)$.
Using the central limit theorem we can approximate the terms.
We have $\sum_i^m y^{(i)}\bm{x_2}^{(i)} \sim \mathcal{N}(0,m\sigma^2)$.
Then, we can approximate
$\frac{1}{k}\sum_j^k \sum_i^m y^{(i)}\bm{x_2}^{(i)}\bm{x_2}^{(j)}$
by $t_1\sim\mathcal{N}(0,\frac{mD}{k}\sigma^4)$ and approximate
$\frac{1}{k}\sum_p^k \sum_i^m y^{(i)}\bm{x_2}^{(i)}\bm{x_2}^{(p)}$
by $t_2\sim\mathcal{N}(0,\frac{mD}{k}\sigma^4)$~\citep{maini2021datasetinference}.
Thus, we get
$t\sim\mathcal{N}(0,\frac{2mD}{k}\sigma^4)$, and

\begin{equation}
\begin{aligned}
     &\quad\ \mathbb{P}[\Psi(\find, \privatedata; \distribution) = 1]
     =
     \mathbb{P}[t\geq\lambda]
     \\&=
     \mathbb{P}[\sqrt\frac{2mD}{k}\sigma^2 Z\geq \lambda]
     =\mathbb{P}[Z\geq\frac{\sqrt{k}\lambda}{\sqrt{2mD}\sigma^2}]
     \\&= 1 - \Phi(\frac{\sqrt{k}\lambda}{\sqrt{2mD}\sigma^2}),
\end{aligned}
\end{equation}
where $Z\sim\mathcal{N}(0,1)$.
The optimal threshold is given as $\lambda = \frac{D\sigma^2}{2}$,
\begin{equation}\label{eq.fp}
     \mathbb{P}[\Psi(\find, \privatedata; \distribution) = 1] = 1-\Phi(\frac{\sqrt{kD}}{2\sqrt{2m}}).
\end{equation}

From Equation~\ref{eq.fp}, we see that the probability of false positives relies on the number of points used for the verification $k$ and the size of $D$.
\oldtext{}{Combining with Lemma~\ref{lemma.2}, the proof is complete.}
\end{proof}

In our subspace where $D$ is bounded for the model with high accuracy, the success of $\di$ is directly related to the number of samples used for the verification.
This is similar to the analysis of failure of membership inference in the original paper when the $k$ is extremely low, e.g. only $10$ samples.
In the $\di$ paper, it was explained
that $\di$ succeeds because it calculates the average margin for multiple verification samples; whereas membership inference fails as it relies on per-sample decision.
So when the number of tested samples is smaller, the success rate of $\di$ will be close to $0.5$, just like for membership inference.
In Figure~\ref{fig:fpr}, we show the probability of an FP (Equation~\ref{eq.fp}) for different values of $k$; even for $k=10000$
the probability is $0.309$
.

Hence, even the simple linear setup, $\Psi(f, \dataset; \distribution)$ has false positives with high probability; in particular, when the fraction of tested samples is small.

\subsection{Non-linear Suspect Models}
\label{sec:nl-fps}

Having demonstrated the limitations of the linear model, we now focus on non-linear suspect models.
The intuition is based on the margin-based generalization bounds.
Note that the generalization bounds states that the expected error of the margin based loss function is bounded, and the bound is mostly related to the distribution~\citep{neyshabur2018pac}.
Since $\di$ assumes all the datasets follow the distribution $\distribution$, our intuition is to directly use the generalization bounds and the triangle inequality to prove the similarity of the models trained on the same distribution.

\subsubsection{Theoretical Motivation}
\label{sec:nl-theory}

Let $f_{\bm{w}}$ be a real-valued classifier $f_{\bm{w}}:\mathcal{X}\rightarrow \mathbb{R}^k$, $||x||\leq B$ with parameters $\bm{w} = \{W_i\}_{i=1}^d$.
For any distribution $\distribution$ and margin $p(f,\bm{x}) = f(\bm{x})[y]-max_{j\neq y}f(\bm{x})[j]\leq\gamma$, where $\gamma>0$.
The margin is the same as for the linear model with labels $y\in\{-1,+1\}$.
Then, we define the margin loss function as:
\begin{equation}
    \mathcal{L}_\gamma(f,y) = \mathbb{P}_{(\bm{x},y)\sim\distribution}[f(x)[y]-max_{j\neq y}f(x)[j]\leq\gamma].
\end{equation}

Note that the PAC-Bayes framework~\citep{neyshabur2018pac} provides guarantees for any classifier $f$ trained on data from a given distribution.
We define the expected loss of a classifier $f$ on distribution $\distribution$ as $\mathcal{L}_\distribution:=E_{(\bm{x},y)\sim\distribution}[\mathcal{L}(f(\bm{x}),y)]$ and the empirical loss on a dataset $\dataset$ as $\hat{\mathcal{L}}_\dataset:=\frac{1}{m}\sum_{(\bm{x},y)\in\dataset}[\mathcal{L}(f(\bm{x}),y)]$.
Then, for a $d-$layer feed-forward network $f$ with parameters $\bm{w}=\{W_i\}_{i=1}^d$ and ReLU activation~\citep{neyshabur2018pac}.
The empirical loss is very close to the expected loss.
For any $\sigma, \gamma>0$, with probability $1-\sigma$ over the training set, we have:

\begin{equation}
    \begin{aligned}
    |\mathcal{L}_\distribution(f_\dataset)- \hat{\mathcal{L}}_\dataset(f_\dataset)|\leq \mathcal{O}(\epsilon),
    \end{aligned}
\end{equation}

where $\epsilon = \sqrt{\frac{B^2d^2h ln(dh) \prod_{i=1}^d||W_i||^2_2\sum_{i=1}^d\frac{||W_i||_F^2}{||W_i||_2^2}+ln\frac{dm}{\sigma}}{\gamma^2m}}$, and $h$ is the upper bound dimension for $\{W_i\}_{i=1}^{d}$.

This PAC-Bayes based generalization guarantee states that for a model $f$,
the distance between the empirical loss and the expected loss is bounded, and the bound can be very small when the model's margin is large.
Thus, we can expect that the margins of $f$ on any dataset that follows a given distribution to be bounded by $O(\epsilon)$.
This contradicts the intuition of $\di$.
$\di$'s intuition is that the difference of margins of two datasets $\privatedata$ and $\independentdata$ on the model $f$ is large enough such that $\di$ is able to distinguish whether the model $f$ is stolen.

Moreover, since $\di$ assumes that $\privatedata$ and $\independentdata$ follow the same distribution $\distribution$,
we can show that the margins for $\fv$ and $\find$ are similar to each other.

\begin{theorem}[k-independent False Positives with Non-linear Suspect Models]\label{thm.2}
For the victim private dataset $\privatedata\sim\distribution$ and an independent dataset $\independentdata\sim\distribution$,
let $f_{\bm{w}}$ be a $d-$layer feed-forward network with ReLU activations and parameters $\bm{w}=\{W_i\}_{i=1}^d$.
Assume that $\fv$ is trained on $\privatedata$ and $\find$ is trained on $\independentdata$, $\fv$ and $\find$ have the same structure. Then, for any $B,d,h,\epsilon>0$ and any $\bm{x}\in\mathcal{X}$, there exist a prior $\mathcal{P}$ on $\bm{w}$, s.t. with probability at least $\frac{1}{2}$,
\begin{equation}
    \begin{aligned}
    |E(p(\fv,\bm{x})- p(\find,\bm{x})) |\leq \epsilon.
    \end{aligned}
\end{equation}
\end{theorem}

The details of the proof are in the Appendix~\ref{app:thm.2}.
Hence, for any two models trained on the same distribution, the expectation of margins for any sample are similar.
Given that $\di$ works by distinguishing the difference of margins for two models, it will result in false positives with probability at least $\frac{1}{2}$ (Theorem~\ref{thm.2}).

\subsubsection{Empirical Evidence}
\label{sec:nl-evidence}

\begin{figure*}[t]
    \resizebox{1.\textwidth}!{
    \begin{tabular}{ccc}
        \includegraphics[]{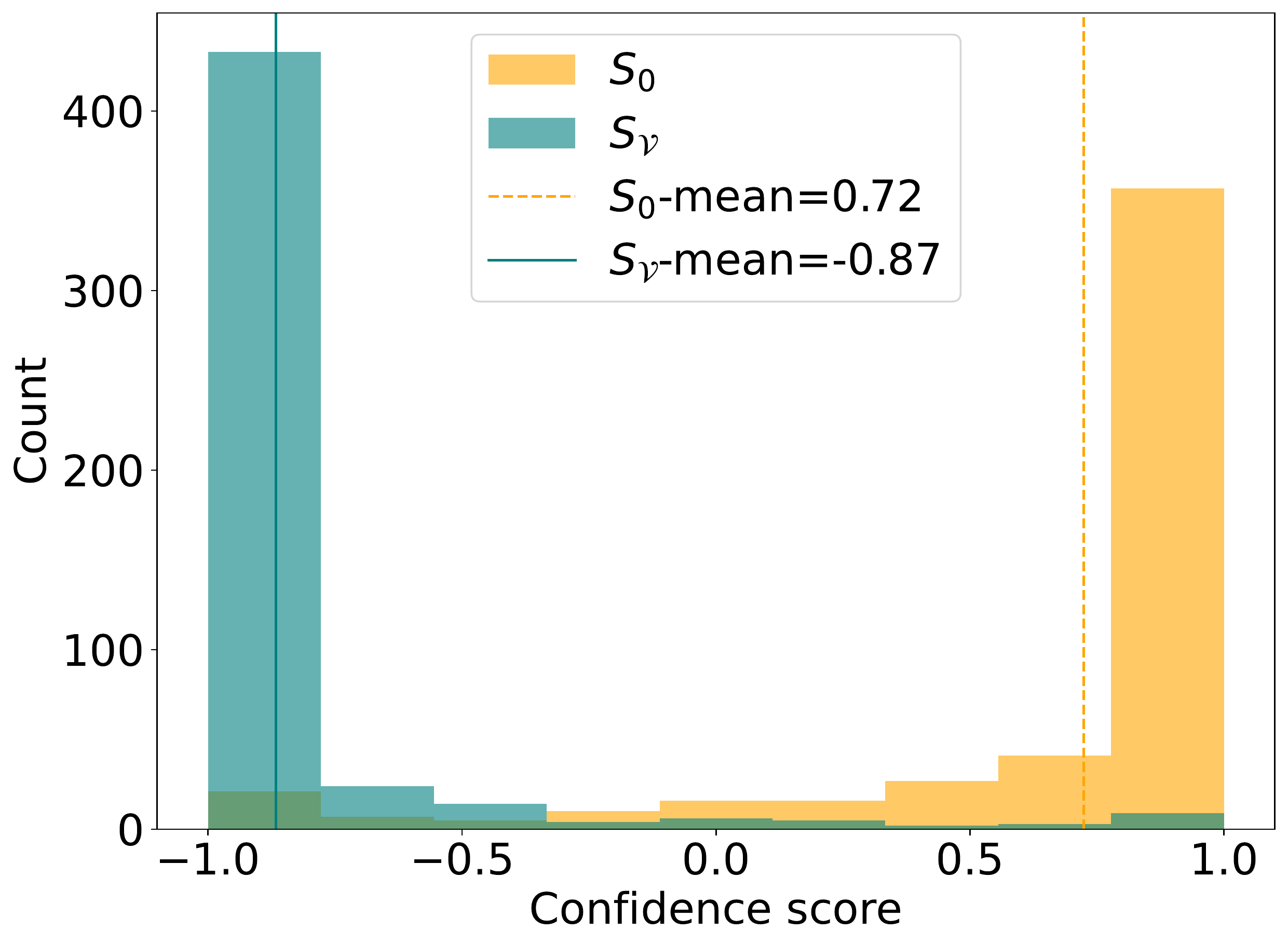} &
        \includegraphics[]{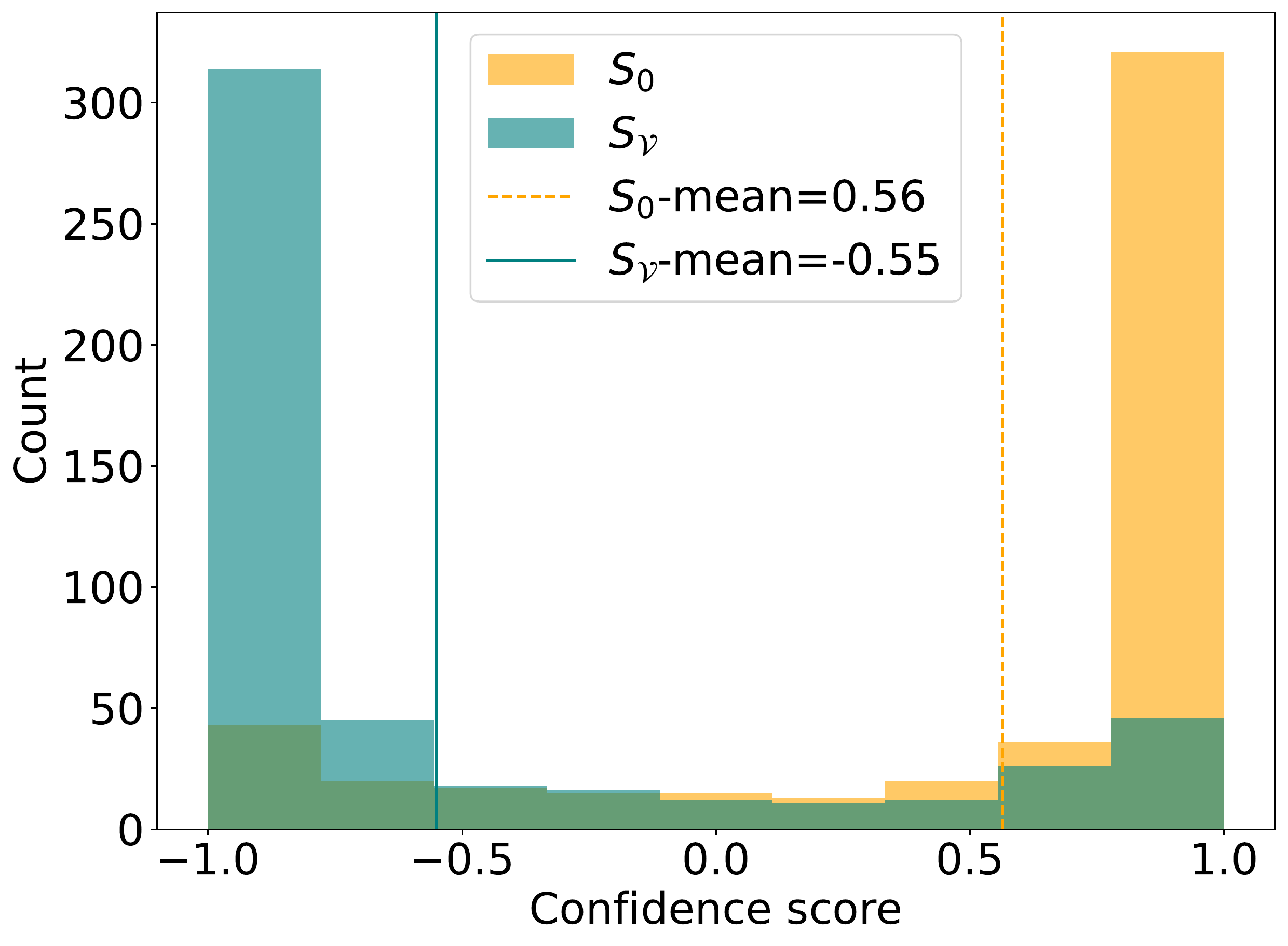} &
        \includegraphics[]{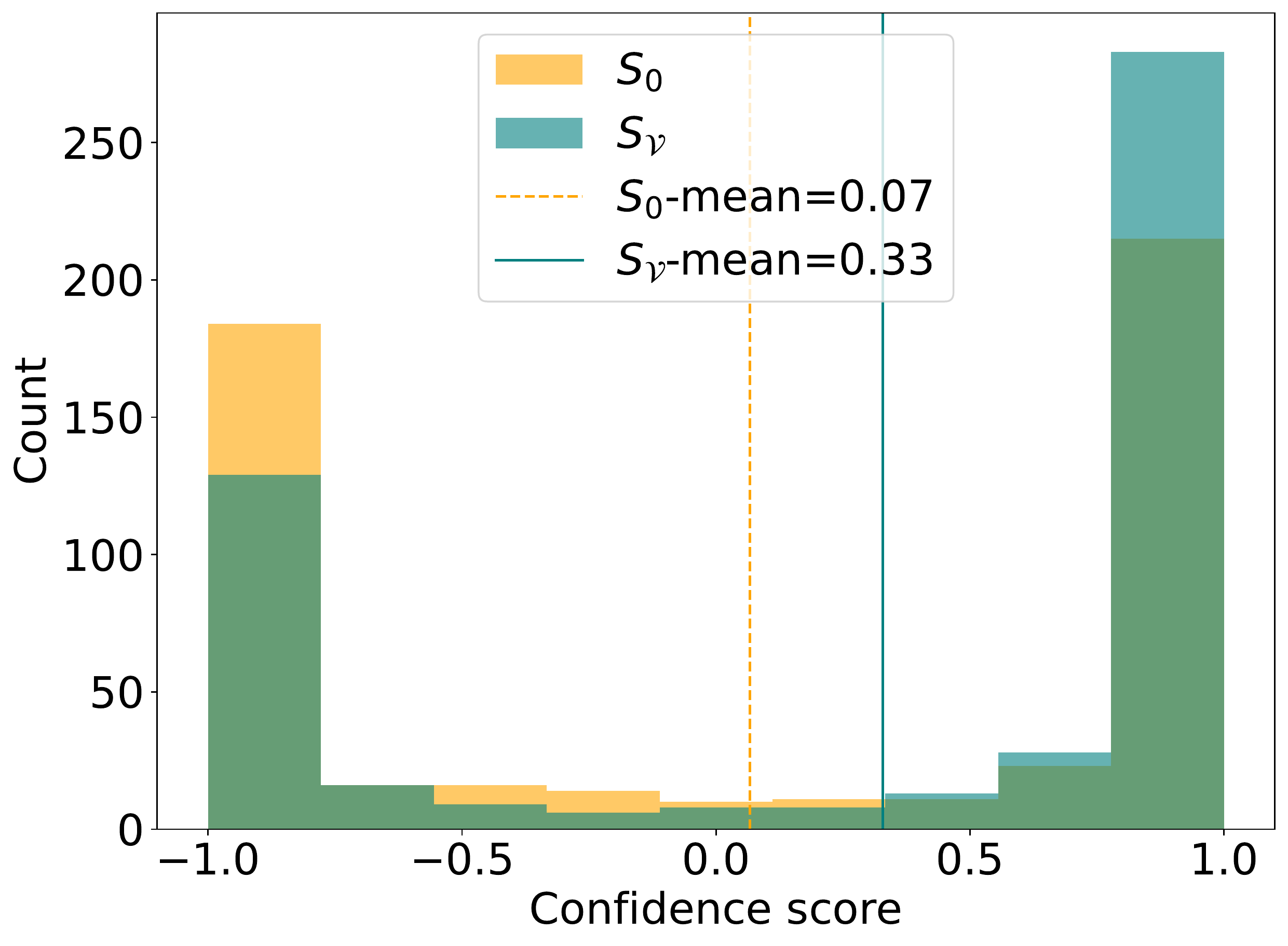} \\
    \end{tabular}}
    \caption{Left to right: $\fv$, $\find$, $\fo$. Comparison of distributions of the confidence scores assigned to the embeddings by $\distinguisher$. $\Delta\mu$ is smaller for $\find$ than for $\fv$ but large enough to trigger an FP.}
    \label{fig:distribution_comp}
\end{figure*}

\begin{figure*}[t]
    \centering
    \resizebox{1.\textwidth}!{
    \begin{tabular}{cc}
        \includegraphics[width=0.5\columnwidth]{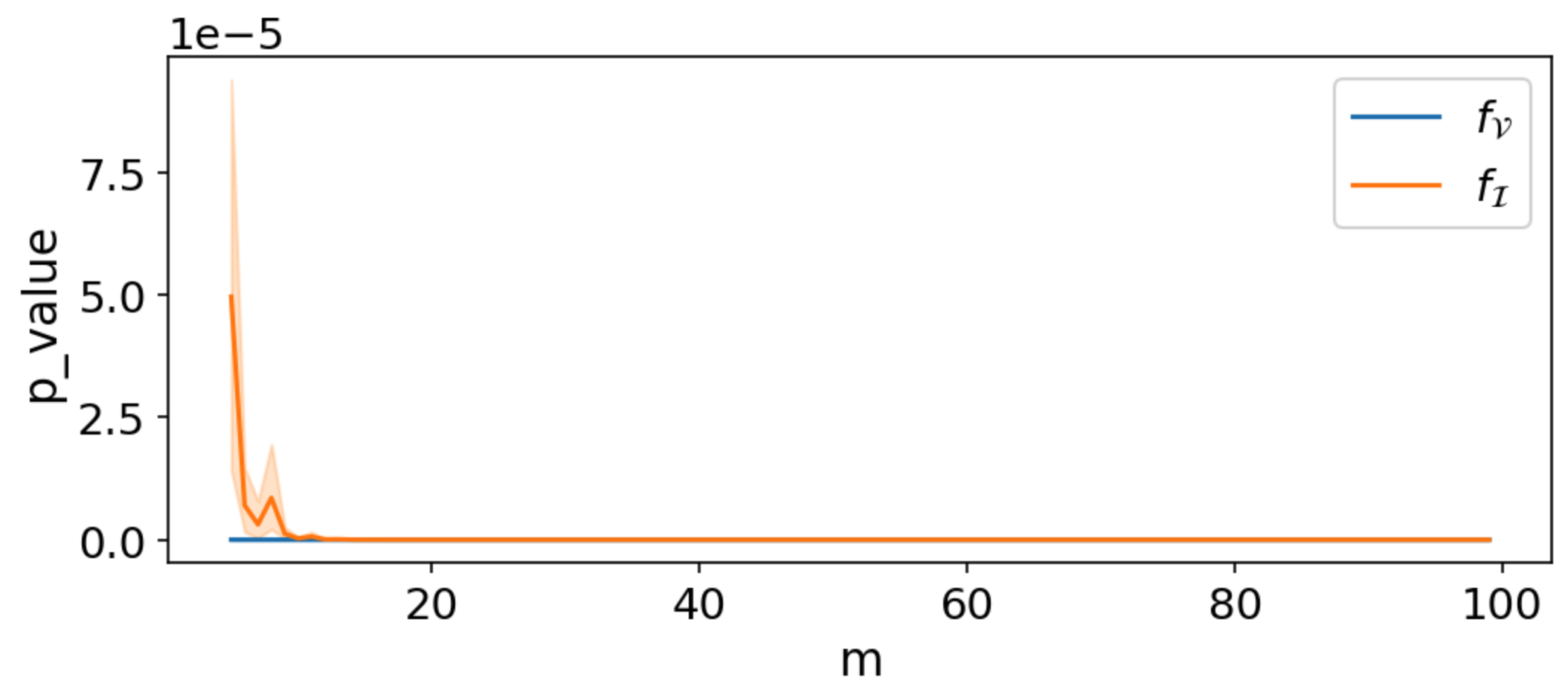} &
        \includegraphics[width=0.5\columnwidth]{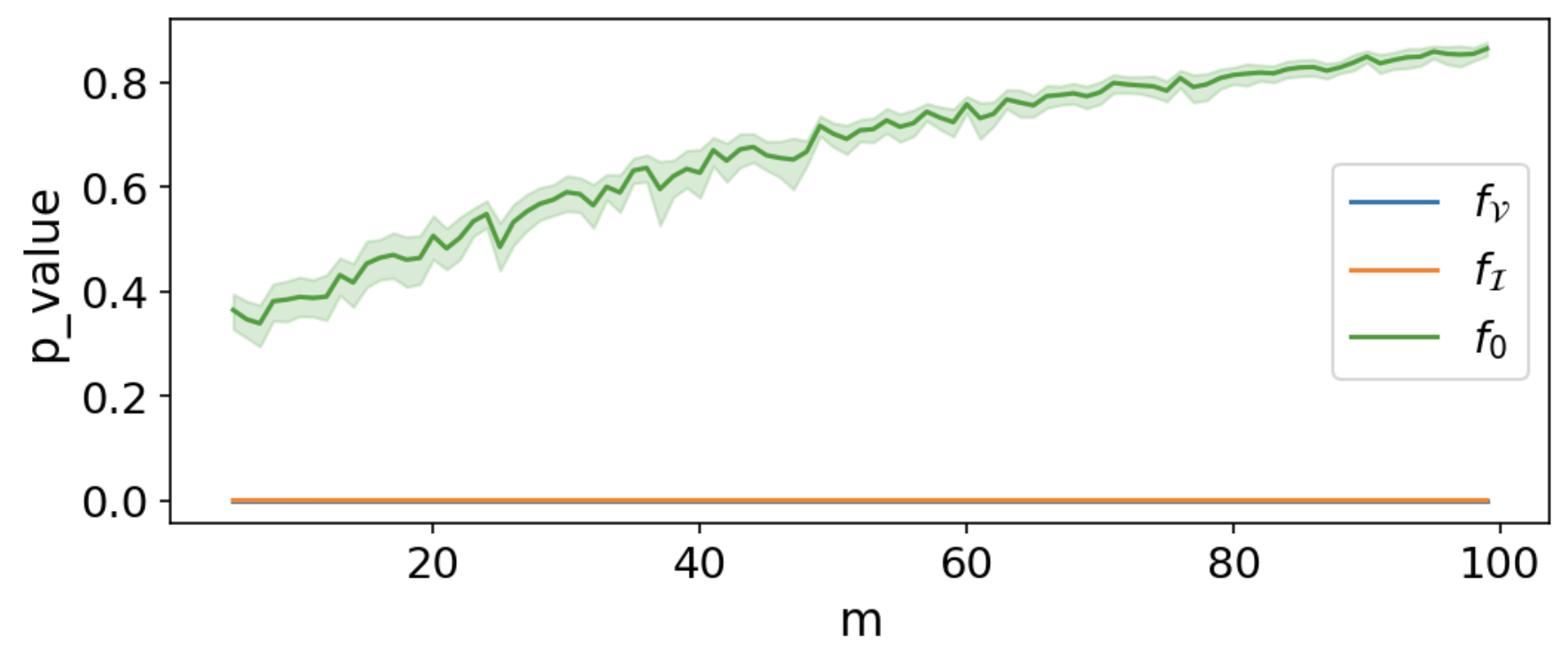} \\
    \end{tabular}}
    \caption{\textbf{Left:} Comparison of the verification confidence of $\fv$ and $\find$. FP becomes stronger (lower p-value) as more samples are revealed. \textbf{Right:} same comparison, however, we include $\fo$ to show the desirable behaviour of an independent model.}
    \label{fig:ver-more-data}
\end{figure*}

Having proved the existence of FPs for non-linear models, we now focus on empirically confirming it.

First, recall the original experiment setup~\citep{maini2021datasetinference}; let us consider the following two models: 1) $\fv$ trained using $\privatedata$, and 2) $\fo$ trained using $\publicdata$.
In the original formulation, e.g. for CIFAR10, CIFAR10-train ($50,000$ samples) is used as $\privatedata$, and CIFAR10-test is used as $\publicdata$ ($10,000$ samples).
Recall that $\victim$ uses their $\privatedata$ and $\publicdata$ to obtain the embeddings that are then used to train the regression model $\distinguisher$.

$\di$ was shown to be effective against several post-processing used to obtain \textit{dependent models} which are expected to be flagged as stolen - true positives
However, the independent model $\fo$ is trained on $\publicdata$ --- the same data that is used to train $\distinguisher$.
This means that the same dataset $\publicdata$ is used both to train $\distinguisher$ and subsequently, to evaluate it.
This is likely to introduce a bias that overestimates the efficacy of $\distinguisher$ and $\di$ as a whole.

To address this, and test whether $\di$ works for a more reasonable data split, we use the following setup:

\begin{enumerate}[label=\textbf{\arabic*)}]
    \item randomly split CIFAR10-train into two subsets ($A_{train}$ and $B_{train}$) of $25,000$ samples each;
    \item assign $\privatedata = A_{train}$, and train $\fv$ using it;
    \item continue using CIFAR10-test as $\publicdata$ (nothing changes), and train $\fo$ using it;
    \item $g_V$ is trained using the embedding for $\publicdata$ and the new $\privatedata$, obtained from the new $\fv$;
    \item assign $\independentdata = B_{train}$, independent data of a third-party $\independent$, who trains their model $\find$.
\end{enumerate}

This way, we have an \textit{independent} model $\find$ that was trained on data from the same distribution $\distribution$ as $\privatedata$ but data that was not seen by $\distinguisher$\footnote{We use the official implementation of $\di$, together with the architectures and training loops. Our changes are limited to the data splits only.}.
We use an analogous split for CIFAR100.

Recall that to determine whether the model is stolen, $\di$ obtains the embeddings for private ($\privatedata$) and public ($\publicdata$) samples.
Then it measures the confidence for each of the embeddings using the regressor $\distinguisher$.
For a model derived from $\victim$'s $\privatedata$, the mean difference ($\Delta\mu$) between the confidence assigned to $\privatedata$ and $\publicdata$ should be large.
If the model is not derived from $\privatedata$, the difference should be small.
The decision is made using the hypothesis test that compares the distributions of measures from $\distinguisher$.

In Figure~\ref{fig:distribution_comp} we visualise the difference in the distributions for three models.
For $\fv$ we observe two separable distributions with a large ($\Delta\mu$), while for $\fo$ the difference is small --- $\di$ is working as intended.
However, for $\find$, even though $\Delta\mu$ is smaller than for $\fv$ it is sufficiently large to reject $H_0$ with high confidence.
Therefore, $\find$ is marked as stolen, a false positive,
In Table~\ref{tab:fps} we provide $\Delta\mu$ and the associated p-values for multiple random splits.
In Figure~\ref{fig:ver-more-data}, we show the results for verification, using Blind Walk, with more data (up to $k=100$ private samples).
As we increase the number of revealed private samples, the confidence of $\di$ increases both for $\fv$ (true positive) and $\find$ (false positive).

\setlength{\extrarowheight}{.3em}
\begin{table}[t]
    \centering
    \caption{Verification of an independent model trained on the same data distribution triggers an FP. Also, we report the accuracy of the models on the test set. We provide the mean and standard deviation computed across five runs. Verification done using $k=10$ private samples. FPs become more significant as $k$ increases. FPs are \bad{highlighted}.}
    \label{tab:fps}
    \small
    \begin{tabular}{|c| c c c| c c c|}\hline
              & \multicolumn{3}{c|}{CIFAR10}                                    & \multicolumn{3}{c|}{CIFAR100}\\
        Model & Accuracy      & $\Delta\mu$         & p-value                   & Accuracy      & $\Delta\mu$         & p-value\\\hline
        \fv   & $0.87\pm0.03$ & $1.62\pm0.08$       & $10^{-18}\pm10^{-18}$     & $0.65\pm0.00$ & $1.80\pm0.04$       & $10^{-30}\pm10^{-30}$\\
        \find & $0.87\pm0.03$ & \bad{$1.14\pm0.12$} & \bad{$10^{-8}\pm10^{-8}$} & $0.66\pm0.01$ & \bad{$0.32\pm0.03$} & \bad{$10^{-2}\pm10^{-2}$}\\
        \fo   & $0.64\pm0.02$ & $-0.29\pm0.12$      & $0.46\pm0.04$             & $0.51\pm0.01$ & $-0.37\pm0.11$      & $0.67\pm0.15$\\ \hline
    \end{tabular}
\end{table}

We discuss the implications of our findings in Section~\ref{sec:data-finger-viability}.

%% file: 4falsenegatives.tex

\section{False Negatives in Dataset Inference}
\label{sec:false-negatives}

Having demonstrated the existence of false positives, we now show that $\di$ can suffer from false negatives (FNs).
$\adv$ can avoid detection by regularising $\fa$, and thus changing the prediction margins.
This in turn, will mislead $\di$ into flagging $\fa$ as independent.

\begin{wrapfigure}[18]{r}{0.5\textwidth}
    \centering
      \includegraphics[width=0.45\textwidth]{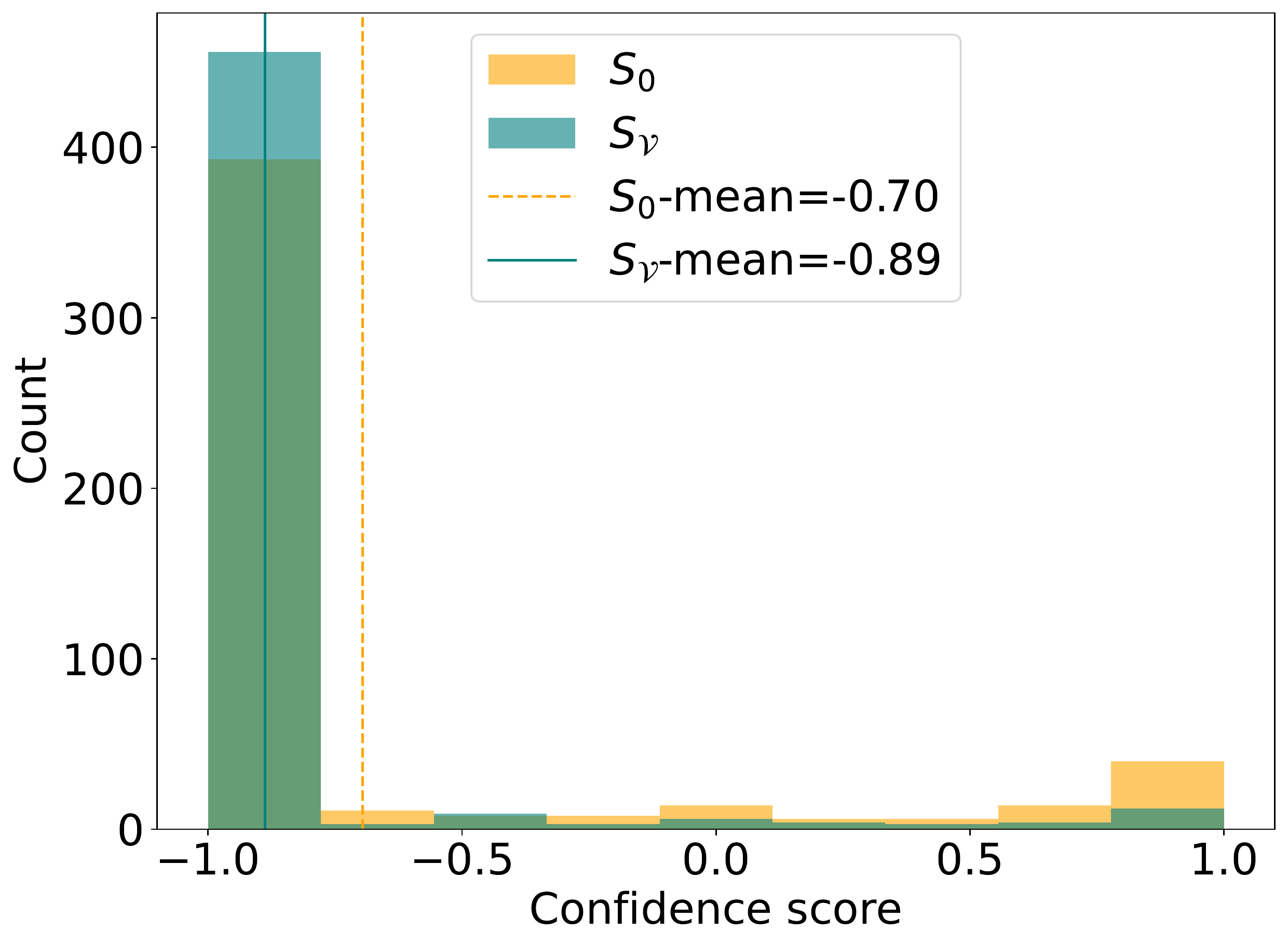}
      \caption{Confidence scores assigned to  embeddings by $\distinguisher$ obtained from $\fa$. $\Delta\mu$ is small enough to trigger FNs.}
    \label{fig:distribution_adv}
  \end{wrapfigure}

Recall that Blind Walk relies on finding the prediction margin by querying perturbed samples designed to cause a misclassification.
In order to avoid detection, $\adv$ needs to make the prediction margin robust to such perturbations.
We do so using adversarial training: a popular regularisation method used to provide robustness against adversarial examples.

The original paper considered three adversary models:
1) $\adv_\mathcal{D}$ who has access to $\victim$'s $\privatedata$;
2) $\adv_\mathcal{Q}$ who has query (i.e. black-box) access to $\fv$, and corresponds to a model extraction attack;
3) $\adv_\mathcal{M}$ who has white-box access to $\fv$.

\underline{$\adv_\mathcal{D}$}: we evaluate adversarial training as a way to avoid detection in the $\adv_\mathcal{D}$ setting -- $\adv_\mathcal{D}$ steals $\victim$'s $\privatedata$ and trains their own model $\fa$.
Also, $\fa$ has the same architecture and hyperparameters as $\fv$, but is adversarially trained
\footnote{We use the official implementation of $\di$, together with the architectures and training loops. Our changes are limited to adding adversarial training.}.

During adversarial training, each training sample $(\vx, y)$ is replaced with an adversarial example that is misclassified $\fa(\vx + \gamma) \ne y$.
There exist many techniques for crafting adversarial examples.
We use \emph{projected gradient descent}~\citep{madry2017towards} (PGD), and we set $\gamma=10/255$ (under $l_\infty$).

In Figure~\ref{fig:distribution_adv} we visualise the difference in the distributions of scores assigned by $\distinguisher$ to $\fa$ embeddings derived for $\privatedata$ and $\publicdata$.
We observe that the distributions are not clearly separable and result in low $\Delta\mu$, and hence $H_0$ cannot be rejected.
Therefore, $\fa$ is marked as an independent model, a false negative.
In Table~\ref{tab:fns} we provide $\Delta\mu$ and the associated p-values for multiple runs.

Note that adversarial training comes with an accuracy trade-off.
In our experiments, the accuracy of $\fa$ goes from $0.92\pm0.01$ to $0.86\pm0.01$ (CIFAR10), and $0.73\pm0.03$ to $0.60\pm0.01$ (CIFAR100).
However, recent work shows that one can adversarially train models without incurring as much accuracy loss~\citep{cui2021learnable,debenedetti2022light,wang2023better}.
We study how the amount of noise affects the verification in Section~\ref{sec:ver-more-noise}.
Also, we discuss the resulting implications in Section~\ref{sec:data-finger-viability}.

We consider $\adv_\mathcal{D}$ in detail because triggering FNs against $\di$ via adversarial training is harder for $\adv_\mathcal{D}$ -- the models have the same architecture, and are trained using exactly the same data.

\underline{$\adv_\mathcal{Q}$}: the same approach also applies in the $\adv_\mathcal{Q}$ setting -- $\adv_\mathcal{Q}$ conducts a model extraction attack with their attack dataset, and can train $\fa$ with adversarial training (without needing access to $\privatedata$).

\underline{$\adv_\mathcal{M}$}: while this approach is not directly applicable to the $\adv_\mathcal{M}$ adversary, an adversary with $\adv_\mathcal{M}$ capabilities and sufficient data (not the same as $\privatedata$) can indeed mount the same attack as $\adv_\mathcal{Q}$ by distilling the model themselves.

\setlength{\extrarowheight}{.3em}
\begin{table}
    \centering
    \caption{$\fa$ adversarially trained on $\privatedata$ results in a false negative. Also, we report the accuracy of the models on the test set. We provide the mean and standard deviation computed across five runs. Verification done using $k=10$ private samples. FNs are are \bad{highlighted}.}
    \label{tab:fns}
    \small
    \begin{tabular}{|c| c c c| c c c|}\hline
              & \multicolumn{3}{c|}{CIFAR10}                                 & \multicolumn{3}{c|}{CIFAR100}\\
        Model & Accuracy      & $\Delta\mu$          & p-value               & Accuracy      & $\Delta\mu$          & p-value\\\hline
        \fv   & $0.92\pm0.01$ & $1.59\pm0.04$        & $10^{-21}\pm10^{-16}$ & $0.73\pm0.03$ & $1.58\pm0.10$        & $10^{-31}\pm{10^31}$\\
        \fa   & $0.86\pm0.01$ & \bad{$0.12\pm 0.06$} & \bad{$0.15\pm0.07$}   & $0.60\pm0.01$ & \bad{$0.11\pm0.02$}  & \bad{$0.43\pm0.18$}\\
        \fo   & $0.64\pm0.02$ & $-0.29\pm0.12$       & $0.46\pm0.04$         & $0.51\pm0.01$ & $-0.37\pm0.11$       & $0.67\pm0.15$\\ \hline
    \end{tabular}
\end{table}

%% file: 5countermeasures.tex

\section{Towards Countermeasures}
\label{sec:countermeasures}

In Sections~\ref{sec:false-positives} and~\ref{sec:false-negatives} we showed that $\di$ suffers from false positives and false negatives in certain settings.
In this section, we explore whether potential countermeasures exist.
First, we analyse possible tweaks to $\distinguisher$ that could help alleviate the FPs (Section~\ref{sec:powerful-gv}).
Next, we test if increasing the amount of noise used during the embedding generation avoids FNs despite adversarial training (Section~\ref{sec:ver-more-noise}).

\subsection{Avoiding FPs via a Modified Distinguisher}
\label{sec:powerful-gv}


Recall that $\di$ does not suffer from any FPs in the original data split where both the $\fo$ and $\distinguisher$ are trained on $\publicdata$.
One possible explanation for the FPs is that $\distinguisher$ overfits to  $\publicdata$, and does not generalise to $\independentdata$.
To verify this hypothesis, we augment $\publicdata$ with a portion of $\independentdata$ (of size $|\publicdata|$), which is then used to train $\distinguisher$.
For $\find$, the p-value still remains low and hence triggers an FP (Table~\ref{tab:fps_Pdis1}).

We then consider an alternative hypothesis --- $\distinguisher$ underfits and fails to predict the margins correctly.
To examine this, we increase the size of $\distinguisher$.
The original $\distinguisher$ is a two-layer network with $tanh$ activation.
We increase the number of layers to four.
Although this does not eliminate the FPs, it does increase the average p-value (Table~\ref{tab:fps_Pdis2}).

In conclusion, neither of these approaches alleviates the FPs. 
The margins of different models trained on the same distribution (estimated by $\distinguisher$) are difficult to distinguish.
Finding a robust approach to avoid FPs in $\di$ remains an open problem.

\setlength{\extrarowheight}{.3em}
\begin{table}[t]
    \caption{Verification of independent models trained on the same data distribution as \fv: a) using more data to train $\distinguisher$; b) using a bigger, four-layer $\distinguisher$ trained with dropout. We provide the mean and standard deviation computed across five runs. Verification done using $k=10$ private samples. FPs are \underline{\textcolor{red}{highlighted}}.}
    \label{tab:fps_Pdis}
    \begin{subtable}[h]{0.45\textwidth}
        \centering
        \small
        \begin{tabular}{|cc|c c|}\hline
            Model & Accuracy      & $\Delta\mu$                    & p-value \\\hline
            \fv   & $0.87\pm0.03$ & $1.34\pm0.08$                  & $10^{-31}\pm10^{-30}$ \\
            \find & $0.87\pm0.03$ & \bad{$0.83\pm0.13$} & \bad{$10^{-11}\pm10^{-10}$} \\
            \fo   & $0.64\pm0.02$ & $0.08\pm0.01$                  & $0.15\pm0.08$ \\ \hline
        \end{tabular}
        \caption{FPs remain when $\distinguisher$ is trained with more data.}
        \label{tab:fps_Pdis1}
    \end{subtable}
    \hfill
    \begin{subtable}[h]{0.45\textwidth}
        \centering
        \small
        \begin{tabular}{|cc|c c|}\hline
            Model & Accuracy      & $\Delta\mu$                    & p-value \\\hline
            \fv   & $0.87\pm0.03$ & $1.30\pm0.17$                  & $10^{-17}\pm10^{-16}$ \\
            \find & $0.87\pm0.03$ & \bad{$0.84\pm0.16$} & \bad{$10^{-7}\pm10^{-6}$} \\
            \fo   & $0.64\pm0.02$ & $0.05\pm0.01$                  & $0.37\pm0.30$ \\ \hline
        \end{tabular}
        \caption{FPs remain when using a bigger $\distinguisher$.}
        \label{tab:fps_Pdis2}
     \end{subtable}
\end{table}

\subsection{Avoiding FNs via Verification with More Noise}
\label{sec:ver-more-noise}

\setlength{\extrarowheight}{.3em}
\begin{table}[h]
    \centering
    \caption{Impact of added noise (maximum number of perturbation steps) during verification on $\di$ success  (\textbf{baseline} $\mathbf{50}$ steps). Using more noise does not prevent FNs against $\fa$ but increases the standard deviation across all experiments, thus negatively impacting verification of $\fo$. We provide the mean and standard deviation computed over five runs. Verification done using $k=10$ private samples. FNs are \bad{highlighted}.}
    \label{tab:noise-added}
    \small
    \begin{tabular}{|c|c c c c| c c c c|}\hline
                     & \multicolumn{4}{c|}{CIFAR10}                                                                  & \multicolumn{4}{c|}{CIFAR100}\\
        Model        & Accuracy                       & Steps         & $\Delta\mu$          & p-value               & Accuracy                       & Steps         & $\Delta\mu$         & p-value\\
        \hline
        \fv          & $0.92\pm0.01$                  & $\mathbf{50}$ & $1.59\pm0.04$        & $10^{-21}\pm10^{-16}$ & $0.73\pm0.03$                  & $\mathbf{50}$ & $1.58\pm0.10$       & $10^{-31}\pm{10^31}$\\
        \hline
\multirow{4}{*}{\fa} & \multirow{4}{*}{$0.86\pm0.01$} & $25$          & \bad{$0.09\pm 0.04$} & \bad{$0.09\pm0.07$}   & \multirow{4}{*}{$0.60\pm0.01$} & $25$          & \bad{$0.12\pm0.02$} & \bad{$0.39\pm0.22$}\\
                     &                                & $\mathbf{50}$ & \bad{$0.12\pm 0.06$} & \bad{$0.15\pm0.07$}   &                                & $\mathbf{50}$ & \bad{$0.11\pm0.02$} & \bad{$0.43\pm0.18$} \\
                     &                                & $100$         & \bad{$0.10\pm 0.05$} & \bad{$0.08\pm0.09$}   &                                & $100$         & \bad{$0.13\pm0.01$} & \bad{$0.50\pm0.16$}\\
                     &                                & $200$         & \bad{$0.14\pm 0.08$} & \bad{$0.16\pm0.11$}   &                                & $200$         & \bad{$0.12\pm0.01$} & \bad{$0.47\pm0.24$}\\
                     \hline
\multirow{2}{*}{\fo} & \multirow{2}{*}{$0.64\pm0.02$} & $\mathbf{50}$ & $-0.29\pm0.12$       & $0.46\pm0.04$         & \multirow{2}{*}{$0.51\pm0.01$} & $\mathbf{50}$ & $-0.37\pm0.11$      & $0.67\pm0.15$\\
                     &                                & $100$         & $-0.19\pm0.16$       & $0.37\pm0.12$         &                                & $100$         & $-0.43\pm0.03$      & $0.72\pm0.09$\\
                     \hline
    \end{tabular}
\end{table}

If $\victim$ suspects that $\adv$ might be using adversarial training to avoid detection, they can carry out the verification with more noise in the attempt to circumvent the effect of adversarially training \fa.

In the experiments presented in Section~\ref{sec:false-negatives}, the average noise added during Blind Walk is $0.12\pm0.05$ (under $\ell_\infty$), and adversarial training is done with $\gamma=10/255(\approx 0.039$).
In this experiment, we vary the number of maximum steps taken by $\victim$, and hence the maximum amount of added noise.
We consider $\{25, 50, 100, 200\}$ steps (baseline $50$ steps) which corresponds to $\{0.10\pm0.03, 0.12\pm0.05, 0.33\pm15, 0.38\pm23\}$ noise added (under $\ell_\infty$) during the verification.
$\victim$ needs to use more noise for the verification of \emph{any} model, hence we also conduct the experiment for $\fo$ (for $\{50, 100\}$ steps).
The goal is to ensure that the increased amount of noise does not have any negative impact on the independent models. 
Table~\ref{tab:noise-added} summarizes our experiment results.
Using more steps does not improve the result against $\fa$ compared to the baseline: 1) for CIFAR10, the standard deviation of the p-value increases; 2) we do not observe any linear relationship between the noise and $\Delta\mu$ or the associated p-value (neither for CIFAR10 or CIFAR100).

On the other hand, for CIFAR10, the confidence of the verification of $\fo$ decreases.
The standard deviations of $\Delta\mu$ and its associated p-value increase.
Although the p-value remains sufficiently high, using more noise has a negative impact on the verification of $\fo$.
On the contrary, for CIFAR100, the confidence of the verification of $\fo$ increases: $\Delta\mu$ is lower, p-value is higher, and standard deviations decrease.

In conclusion, increasing the amount of noise during Blind Walk does not allow $\victim$ to circumvent $\adv$'s adversarial training. Hence, $\di$ remains susceptible to false negatives induced by adversarial training.

%% file: 6discussion.tex

\section{Discussion}
\label{sec:discussion}

\textbf{Revealing private data.}\label{sec:revealing-samples}
We have shown that $\di$ requires revealing significantly more than $50$ samples to avoid false positives in the case of linear models (Figure~\ref{fig:fpr}).
Since the core assumption of $\di$ is that $\privatedata$ is private, revealing too much of $\privatedata$ during the ownership verification constitutes a privacy threat.
In neither of the settings described in Section 5 of the original DI paper the victim \emph{cannot query the model sufficiently} without leaking the query data to the adversary.
Additionally, it was shown that using more samples gives $\victim$ more information about the prediction margin than using stronger embedding methods~\citep{maini2021datasetinference}.
Model owners that operate in sensitive domains such as healthcare or insurance industry need to comply with strict data protection laws, and hence need to minimise the disclosure.

One potential way to protect the privacy of the private samples used for $\di$ ownership verification is to use oblivious inference~\citep{jianliu2017minionn,juvekar2018gazelle}.
This way $\victim$ could query $\fa$ without revealing $\privatedata$.
Despite recent advances in efficient oblivious inference~\citep{samragh2021oblivbinary,watson2022piranha,samaradzic2021f1,samaradzic2022craterlake}, it requires \emph{all} parties (including $\adv$!) to update their software stacks which may not always be realistic.

\textbf{Viability of ownership verification using training data.}\label{sec:data-finger-viability}
We have demonstrated that $\di$ suffers from FPs when faced with an independent model trained on the same distribution.
While it is reasonable to assume that $\victim$'s data is private, the uniqueness of the distribution is difficult to guarantee in practice.
For example, two model builders may have data from the same distribution because they purchased their training data from a vendor that generates per-client synthetic data from the same distribution (e.g., regional financial data).
In fact, two model builders working on the same narrow domain and independently building models that are intended to represent the same phenomenon, may very well end up using data from the same distribution.

There are other methods that attempt to detect stolen models based on the dataset used to train them~\citep{sablayrolles2020radioactive,pan2022metav}.
However, they rely on flaws in the model to establish the ownership (susceptibility to adversarial examples~\citep{sablayrolles2020radioactive} or membership inference attacks~\citep{pan2022metav}).
Intuitively, given a perfect membership inference attack, a fingerprinting scheme should be possible.
However, recent work shows that for a balanced dataset, only a fraction of records is vulnerable to a confident membership inference attack~\citep{carlini2022lira,duddu2021shapr} which in turn reduces the capabilities of a membership inference-based fingerprinting scheme.
Therefore, any improvements to generalisation or robustness (such as adversarial training or purification~\citep{nie2022DiffPure}) of ML models reduce the surface for ownership verification schemes.

Nevertheless, leveraging advancements in membership inference might make $\di$ more robust, e.g. regularising $\di$ with many shadow models to avoid FPs, or using MIAs that were shown to be more effective against adversarially trained models that overfit~\citep{song2019miaadversarial,song2019miaadversarialconf,hayes2020tradeoffs}. 
However, it is worth noting that the orignal $\di$ paper~\citep{maini2021datasetinference} differentiates $\di$ from membership inference.
$\di$ relies on weak signals over many records unlike MIA that needs strong signal for individual records.
Hence, relying on ideas from MIAs introduces additional considerations, e.g. MIA evasion.

\textbf{False positives in other schemes.}\label{sec:fps-other-schemes}
Our work raises the question whether other watermarking/fingerprinting schemes are also vulnerable. 
Existing schemes for model ownership verification typically focus on robustness against adversaries who want to evade detection.
Systematically examining such schemes for false positives can be an interesting line of future work.
An interesting question is whether a malicious model owner can successfully accuse independent models as stolen by systematically inducing false positives.

\textbf{White-box theft.}\label{sec:whitebox}
Our experiments in Section~\ref{sec:false-negatives} are limited to $\adv$ that trains their own model --- they either steal the data or conduct a model extraction attack.
If $\adv$ obtains an exact copy of the model, they might lack the data to fine-tune it with adversarial training.
Hence, our findings do not apply to the white-box setting.
We leave the examination of other threat models out as future work.

\textbf{Black-box vs. white-box verification setting.}
Our evaluation is focused on the black-box $\di$ setting.
We do not consider the white-box $\di$ setting which uses MinGD.
While white-box $\di$ is feasible in a scenario where  $\victim$ takes $\adv$ (the holder of a suspect model) to court, requiring $\adv$ to provide white-box access to the suspect model, prosecution is an expensive undertaking.
Realistically $\victim$ is likely to first conduct black-box $\di$ to decide whether the expense of prosecution is justified. Therefore, FPs in the black-box $\di$ setting can cause substantial monetary loss to $\victim$.

%% file: 8conclusion.tex

\section{Conclusion}
\label{sec:conclusion}

We analyzed \emph{Dataset Inference} ($\di$)~\cite{maini2021datasetinference}, a promising fingerprinting scheme, to show theoretically and empirically that $\di$ is prone to false positives in the case of independent models trained from distinct datasets drawn from the same distribution.
This limits the applicability of $\di$ only to settings where a model builder uses a dataset with a definitively unique distribution.
Alleviating FPs in other settings may not be possible. 
Furthermore, $\di$ might not be applicable in settings where privacy of the records must be respected, e.g. healthcare.
We also showed that an attacker can use adversarial training to regularise the decision boundaries of a stolen model to evade detection by $\di$ at the cost of a drop in accuracy ($6-13pp$).
Such decrease is acceptable if the model can be further fine-tuned with additional data and recover the performance.

Nevertheless, $\di$ is a promising ML fingerprinting scheme.
Model owners can use our results to make informed decisions as to whether $\di$ is appropriate for their particular settings.

%% file: 9appendix.tex




\input{7related}

\section{Existence of False Positives in Dataset Inference}
\label{app:fps-more-theory}

\textbf{Calculating the prediction margin.}
We assume that the model weights are initialized to zero.
For each sample $x$ in a dataset $\dataset \sim \distribution = \{(\bm{x^{(i)}},y^{(i)})| i=1,...,m\}, y\sim\{-1,+1\}$.
The learning algorithm observes all samples in $S$ once and maximizes the loss function $L(\bm{x},y)=y\cdot f(\bm{x})$.
For the learning rate $\alpha = 1$, the weights are updates as:
\begin{equation}
    w=w+\alpha y^{(i)}\bm{x^{(i)}}.
\end{equation}
Recall that $\bm{x}=(\bm{x_1},\bm{x_2})\in\mathbb{R}^{K+D}$, the weights of the linear model are $\bm{w_1}=m\bm{u}$ and $\bm{w_2}=\sum_{i=1}^{m} y^{(i)}\bm{x_2}^{(i)}$ when the training is completed.

When writing out the linear classifier explicitly, we can easily calculate the prediction margin of each sample $(x,y)$ in $\dataset$,
\begin{equation}
    y\cdot f(\bm{x})= y\cdot(\bm{w_1}\bm{x_1}+\bm{w_2}\bm{x_2})=y\cdot(m\bm{u}\cdot y\bm{u} +\sum_{i=1}^{m}y^{(i)}\bm{x_2}^{(i)}\cdot \bm{x_2}) = c+y\sum_{i=1}^{m}y^{(i)}\bm{x_2}^{(i)}\cdot \bm{x_2}.
\end{equation}

The expectations of the prediction margin for the points in training set $\dataset^{ + } = \{(x,1)|(x,1) \in \dataset\}$ is,
\begin{equation}\label{equ.3}
\begin{aligned}
        E_{\dataset^+}[yf(\bm{x})]&=yc+E_{\dataset^+}[\sum_{i=1}^{m}y^{(i)}\bm{x_2}^{(i)}\cdot \bm{x_2}] = yc+E_{\dataset^+}[\sum_{\bm{x_2}\neq \bm{x_2}^{(i)}}y^{(i)}\bm{x_2}^{(i)}\cdot \bm{x_2} +y\bm{x_2}^2
        ]\\&=c+0+D\sigma^2.
\end{aligned}
\end{equation}

Note that in Equation~\ref{equ.3}, since $\bm{x_2} \sim N(0,D\sigma^2)$, then $\bm{x_2}^2 \sim \chi^2$, $E[\bm{x_2}^{(i)}]=D\sigma^2$.

Consider a new dataset $\publicdata\sim \distribution$, the expectations of the prediction  margin for the points in $\publicdata^+$ are,
\begin{equation}
\begin{aligned}
        E_{\publicdata^+}[yf(\bm{x})]=yc+E_{\publicdata^+}[\sum_{i=1}^{m}y^{(i)}\bm{x_2}^{(i)}\cdot \bm{x_2}]=c.
\end{aligned}
\end{equation}
Finally, we see that the difference of prediction margin of training set $S$ and test set $\publicdata$ is
\begin{equation}
    E_{\dataset^+}[yf(\bm{x})]- E_{\publicdata^+}[yf(\bm{x})] = D\sigma^2.
\end{equation}

\textbf{$\di$'s decision function.}
From the above analysis, we know that the statistical difference between the distribution of training and test data is $D\sigma^2$ which is usually larger than $1$ in numerical.
DI utilizes this difference to predict if a potential adversary's model stole their knowledge.

Since we know that $E_{\publicdata}[yf(\bm{x})]=c$ and $E_{\dataset}[yf(\bm{x})]=c+D\sigma^2$.
Let $\Psi(f, \dataset; \distribution)$ represent the dataset inference victim's decision function.
It is defined as,
\begin{equation}
\Psi(f, \dataset; \distribution) =   \left\{
\begin{aligned}
&1,\ if\ E_{(x,y)\in\dataset}[y\cdot f(\bm{x})]- E_\distribution[y\cdot f(\bm{x})]\geq \lambda,\\
&0,\ otherwise,
\end{aligned}
\right.
\end{equation}
where $\lambda\in [0,D\sigma^2]$ is some threshold that the decision function uses to maximise true positives and minimise false positives.

\textbf{Proof for Lemma~\ref{lemma.2}}\label{app:lem.1} For a linear model $f$ trained on distribution $\distribution$ where $\bm{x} =(\bm{x_1},\bm{x_2})$, $\bm{x_1} = y\bm{u}$,$\bm{x_2}\sim\mathcal{N}(0,\sigma^2)$ and $||\bm{u}||_2\leq \frac{1}{\sqrt{m}}$, $f$ is expected to achieve high accuracy on any sample $(\bm{x},y)$ sampled randomly from $\distribution$ which is independent of the training data set of $f$.

\begin{proof}
Given a linear model $f$ trained on dataset $\dataset \sim \distribution = \{(\bm{x^{(i)}},y^{(i)})| i=1,...,m\}$, and a test sample $(\bm{x},y)$ sampled randomly from $\distribution$ which is independent of $\dataset$, the probability that $(\bm{x},y)$ is correctly classified by $f$ can be represented as:
\begin{equation}
\begin{aligned}
    \mathbb{P}[yf(x)\geq 0 ] &=\mathbb{P}[ m\bm{u}^2 +y\sum_i^my{(i)}\bm{x_2}^{(i)}\bm{x_2}\geq 0 ]
    \\&=\mathbb{P}[y\sum_i^my{(i)}\bm{x_2}^{(i)}\bm{x_2}\geq-m\bm{u}^2]
    \\& \oldtext{}{\leq \mathbb{P}[y\sum_i^my{(i)}\bm{x_2}^{(i)}\bm{x_2}\geq-1]}
\end{aligned}
\end{equation}
Since $\bm{x_2}\sim\mathcal{N}(0,\sigma^2)$ are $D$-dimensional vectors, we can use the central limit theorem to approximate the term. Thus, the internal term can be approximated by a variable $t \sim \mathcal{N}(0,mD\sigma^4)$. Let $Z\sim\mathcal{N}(0,1)$,

\begin{equation}
    \mathbb{P}[yf(x)\geq 0 ] \leq \mathbb{P}[\frac{\sqrt{mD}\sigma^2}{m\bm{u}^2}Z\geq -1 ] = 1- \Phi(-\frac{m\bm{u}^2}{\sqrt{mD}\sigma^2})
\end{equation}
where $\Phi$ is the normal CDF.

\oldtext{}{
For a distribution where the randomness $\sigma^2\geq\frac{1}{\sqrt{m}}\geq\frac{1}{4\sqrt{m}}$.
\begin{equation}
     \mathbb{P}[yf(x)\geq 0 ] \leq 1- \Phi(-\frac{4}{\sqrt{D}}),
\end{equation}
where $\Phi(-\frac{4}{\sqrt{D}})\approx 0.10$.
The linear model $f$ can correctly classify a sample with a probability more than 0.9 only if $D<10$.
}

We can also calculate the accuracy of the training set $\dataset$ similarly. For a training sample  $(\bm{x},y)$ sampled randomly from $\dataset$,
\begin{equation}
\begin{aligned}
    \mathbb{P}[yf(x)\geq 0 ] &=\mathbb{P}[ m\bm{u}^2 +y\sum_i^my{(i)}\bm{x_2}^{(i)}\bm{x_2}\geq 0 ]
    \\&=\mathbb{P}[m\bm{u}^2 +  y^2\bm{x_2}^2 + y\sum_i^{m-1}y{(i)}\bm{x_2}^{(i)}\bm{x_2}\geq 0]
    \\&=\mathbb{P}[ y^2\bm{x_2}^2+y\sum_i^my{(i)}\bm{x_2}^{(i)}\bm{x_2}\geq-m\bm{u}^2]
    \\&\leq \mathbb{P}[ y^2\bm{x_2}^2+y\sum_i^my{(i)}\bm{x_2}^{(i)}\bm{x_2}\geq-1].
\end{aligned}
\end{equation}
Since $ y^2\bm{x_2}^2\geq0$ for any sample in $\dataset$, we have
\begin{equation}
    y^2\bm{x_2}^2+y\sum_i^my{(i)}\bm{x_2}^{(i)}\bm{x_2}\geq y\sum_i^my{(i)}\bm{x_2}^{(i)}\bm{x_2}.
\end{equation}
Then, $ \mathbb{P}_{(\bm{x},y)\in\dataset}\geq \mathbb{P}_{(\bm{x},y)\in\distribution/\dataset}$.
This completes the proof.
\end{proof}

\textbf{Proof for Theorem~\ref{thm.2}}\label{app:thm.2}
Let $f_{\bm{w}}$ be a $d$-layer feed-forward model trained on distribution $\distribution$ with parameters $\bm{w} = \{W_i\}_{i=1}^d$ and the ReLU
activation function.
Assuming a training dataset $\dataset \sim \distribution$, the model is given as $f_{\dataset} = f_{\bm{w}+\bm{u}_{\dataset}}$,
where $\bm{u_{\dataset}}$ is a random variable whose distribution may also depend on $\dataset$.

Since the key to analyze the margin is the output of the model,
we first introduce Lemma~\ref{lemma.per_b} that analyzes the perturbation bound of the model trained on $\dataset$ and $\distribution$.

\begin{lemma}[Perturbation Bound (Lemma 2) in \cite{neyshabur2018pac}]\label{lemma.per_b}
For any $B, d>0$, let $f_{\bm{w}}: \mathcal{X}\rightarrow \mathbb{R}^k$ be a $d-$layer neural network with ReLU activations. Then for any $\bm{w}$, and $\bm{x}\in\mathcal{X}$, and any perturbation $\bm{u}_{\dataset} = \{U_i\}_{i=1}^d$ such that $||U_i||_2\leq\frac{1}{d}||W_i||_2$, the change in the output of the network can be bounded as follow,
\begin{equation}
    |f_{\bm{w}+\bm{u}_{\dataset}}(\bm{x})- f_{\bm{w}}(\bm{x})|\leq
    eB(\prod_{i=1}^d ||W_i||_2)\sum_{i=1}^d\frac{||U_i||_2}{||W_i||_2}.
\end{equation}
\end{lemma}
Since our proof is also based on Lemma~\ref{lemma.per_b}, it is analogous to the analysis of generalization bound in \cite{neyshabur2018pac} and is essentially the same for the first part.
\begin{proof}
The proof involves two parts.
In the first part, we show the maximum allowed perturbation of parameters as shown in~\cite{neyshabur2018pac}.
In the second part, we show that the margin difference of the models trained on $\privatedata$ and $\independentdata$ is also bounded by the perturbation of parameters.
Let $\beta = (\prod_{i=1}^d ||W_i||_2)^{\frac{1}{d}}$, and consider a network with normalized weights $\Tilde{W}_i=\frac{\beta}{||W_i||_2}W_i$.
Due to the homogeneity of the ReLU, we have $f_{\Tilde{\bm{w}}} = f_{\bm{w}}$.
We can also verify that $(\prod_{i=1}^d ||W_i||_2)=\prod_{i=1}^d ||\Tilde{W}_i||_2$ and $\frac{||W_i||_F}{||W_i||_2} = \frac{||\Tilde{W}_i||_F}{||\Tilde{W}    _i||_2}$.
Therefore, it is sufficient to prove the Theorem only for the normalized weights $\Tilde{\bm{w}}$, and hence w.l.o.g we assume that for any layer $i$, $||W_i||_2=\beta$.

Choose the distribution $\mathcal{P}$ of the prior of $\bm{w}$ to be $\mathcal{N}(0,\sigma^2I)$, and consider the random perturbation $\bm{u}_{\dataset}\sim \mathcal{N}(0,\sigma^2I) = \{U_i\}_{i=1}^d$.
Since the prior cannot depend on the learned model $\bm{w}$ or its norm, we set $\sigma$ based on the approximation $\Tilde{\beta}$.
For each value of $\Tilde{\beta}$ on a pre-determined grid, we compute the PAC-Bayes bound, establishing the generalization guarantee for all $\bm{w}$ for which $|\Tilde{\beta}-\beta|\leq\frac{1}{d}\beta$, and ensuring that each relevant value of $\beta$ is covered by some $\Tilde{\beta}$ on the grid.
We then take a union bound over all $\Tilde{\beta}$ on the grid.
For now, we consider a fixed $\Tilde{\beta}$ and the $\bm{w}$ for which $|\beta-\Tilde{\beta}|\leq\frac{1}{d}\beta$, and hence $\frac{1}{e}\beta^{d-1}\leq \Tilde{\beta}^{d-1}\leq e\beta^{d-1}$.

Since $\bm{u}_{\dataset}\sim \mathcal{N}(0,\sigma^2I)$, we get the following bound for the spectral norm of $U_i$~\cite{tropp2012user}:
\begin{equation}
    \mathbb{P}_{U_i\sim\mathcal{N}(0,\sigma^2I)}[||U_i||_2>t]\leq 2he^{-t^2/2h\sigma^2}.
\end{equation}

Taking a union bound over the layers, we get that with probability at least $\frac{1}{\sqrt{2}}$, the spectral norm of perturbation of $U_i$ in each layer is bounded by $\sigma\sqrt{2hln(2dh)}$.
Plugging this spectral norm bound into Lemma~\ref{lemma.per_b} we have that with probability at least $\frac{1}{\sqrt{2}}$ the maximum allowed perturbation bound is:

\begin{equation}
\begin{aligned}
     max_{\bm{x}\in\mathcal{X}}|f_{\bm{w}+\bm{u}_\dataset}(\bm{x})- f_{\bm{w}}(\bm{x})|\leq &eB\beta^d\sum_i\frac{||U_i||_2}{\beta}
     \leq e^2dB\Tilde{\beta}^{d-1}\sigma\sqrt{2hln(2dh)}\leq\frac{\epsilon}{4},
\end{aligned}
\end{equation}
where $\sigma = \frac{\epsilon}{42dB\Tilde{\beta}^{d-1}\sigma\sqrt{2hln(2dh)}}$.
Then we can compute the difference of expectation margins for $\fv$ which is trained on $\privatedata$ and $\find$ which is trained on $\independentdata$. Firstly, we compute the difference margins for any model $f_\dataset$ trained on $\dataset \sim \mathcal{D}$ and the target model $f_{\distribution}$. For any verified dataset $\hat{S}\in \mathcal{D}$,
\begin{equation}
    \begin{aligned}
    &|E (p(f_{\dataset}, \bm{x})) - E(p(f_{\distribution},\bm{x}))|\\=
    &|
    E(f_{\bm{w}+\bm{u}_\dataset}(\bm{x})[y]-max_{j\neq y} f_{\bm{w}+\bm{u}_\dataset}(\bm{x})[j]) -
        E(f_{\bm{w}}(\bm{x})[y]-max_{j\neq y} f_{\bm{w}}(\bm{x})[j])
    |
    \\=&|
    (E (f_{\bm{w}+\bm{u}_\dataset}(\bm{x})[y])-E (f_{\bm{w}}(\bm{x})[y]))-
    (E (max_{j\neq y} f_{\bm{w}+\bm{u}_\dataset}(\bm{x})[j]) -E (max_{j\neq y}f_{\bm{w}}(\bm{x})[j]))
    |\\\leq
    & max_{\bm{x}\in\mathcal{X}}(f_{\bm{w}+\bm{u}_\dataset}(\bm{x})[y]- f_{\bm{w}}(\bm{x})[y]) + max_{\bm{x}\in\mathcal{X}}( max_{j\neq y} f_{\bm{w}+\bm{u}_\dataset}(\bm{x})[j] - max_{j\neq y}f_{\bm{w}}(\bm{x})[j] )
    \\\leq& 2max_{\bm{x}\in\mathcal{X}}|f_{\bm{w}+\bm{u}_\dataset}(\bm{x})- f_{\bm{w}}(\bm{x})|\leq\frac{\epsilon}{2}.
    \end{aligned}
\end{equation}

So, for $\fv$ trained on $\privatedata$ and $\find$ trained on $\independentdata$, we have with probability at least $\frac{1}{2}$ that the predictions margins are bounded by $\epsilon$:
\begin{equation}
    \begin{aligned}
    &|E (p(\fv, \bm{x})) -E (p(\find,\bm{x}))|
    \\\leq& |E(p(\fv, \bm{x})) -E( p(f_{\distribution},\bm{x}))|
    +|E(p(\find, \bm{x})) -E (p(f_{\distribution}),\bm{x})|
    \\\leq& \epsilon.
    \end{aligned}
\end{equation}
\end{proof}

%% file: 7related.tex

\section{Related Work}
\label{sec:related}

\textbf{Model extraction detection and prevention.}
Detection methods rely on the fact that many extraction attacks have querying patterns that are distinguishable from the benign ones~\citep{juuti2019prada,atli2020boogeyman,zheng2020blp,quiring2018forgottensib}.
All of these can be circumvented by the adversary who has access to natural data from the same domain as the victim model~\citep{atli2020boogeyman}.
Prevention techniques aim to slow down the attack by injecting the noise into the prediction, designed to corrupt the training of the stolen model~\citep{orekondy20prediction,lee2018defending,mazeika2022steeringnoise}, or by making all clients participate in consensus-based cryptographic protocols~\citep{dziedzic2022increasing}.
Even though they increase the cost of the attack, they do not stop a determined attacker from stealing the model.

\textbf{Ownership verification.}
There exist many watermarking schemes for neural networks (e.g.~\citep{zhang2018protecting,uchida2017embedding,adi2018turning}) that have the same goal as $\di$ does.
However, they mostly rely on embedding a \emph{backdoor} into a model that serves as a watermark.
Unfortunately, they do not protect against model extraction attacks, and hence, several schemes were proposed that specifically target model extraction~\citep{szyller2021dawn,jia2021entangled}.

It was also shown that adversarial examples~\citep{lukas2021conferrable} can be used to fingerprint a model by finding examples that transfer to models \emph{only} derived from the original.
One could also use adversarial examples to watermark datasets by including some in the training set~\citep{sablayrolles2020radioactive}.
However, adversarial training can be used to weaken both schemes~\citep{lukas2021conferrable,szyller22conflicts}.
On the other hand, if a model is sufficiently vulnerable to membership inference attacks, it can be used to fingerprint it~\citep{pan2022metav}.

Alternatively, one could prove ownership of the model using a proof of the integrity of training (Proof-of-Learning)~\citep{jia2021pol}.
Proof-of-learning relies on the fact that successive training checkpoints can serve as a proof of expended compute, and (probabilistic) correctness of the
training procedure.

We believe that the following work is the closest to ours.
~\cite{lukas2021wmsok} showed that all watermarking schemes are brittle.
They designed various attacks that successfully remove watermarks, or prevent their embedding.
However, they do not evaluate any fingerprinting schemes.
~\cite{zhang2022polspoof} showed that $\adv$ can construct a \emph{spoof} against Proof-of-Learning.
$\adv$ iteratively creates the spoof by finding adversarial examples that match the successive training steps, hence resulting in a false positive.